\documentclass[letterpaper]{article} 
\usepackage{aaai23}
\usepackage{times}  
\usepackage{helvet}  
\usepackage{courier}  
\usepackage[hyphens]{url}  
\usepackage{graphicx} 
\usepackage{natbib}  
\usepackage{caption} 
\DeclareCaptionStyle{ruled}{labelfont=normalfont,labelsep=colon,strut=off} 
\usepackage{algorithm}
\usepackage{algorithmic}
\usepackage{setspace}
\usepackage{subfigure}
\usepackage{colortbl} 
\usepackage{color}

\definecolor{mygray}{gray}{.95}

\usepackage{newfloat}
\usepackage{listings}
\floatstyle{ruled}
\newfloat{listing}{tb}{lst}{}
\floatname{listing}{Listing}

\usepackage{amssymb}
\usepackage{amsmath}
\usepackage{multirow}

\title{From Indoor To Outdoor: Unsupervised Domain Adaptive Gait Recognition}
\author {
	Likai Wang,\textsuperscript{\rm 1}
	Ruize Han, \textsuperscript{\rm 1}
	Wei Feng, \textsuperscript{\rm 1}
	Song Wang \textsuperscript{\rm 2}
}
\affiliations {
	\textsuperscript{\rm 1} College of Intelligence and Computing, Tianjin University\\
	\textsuperscript{\rm 2} University of South Carolina Columbia, USA\\
	\{kkww, han\_ruize\}@tju.edu.cn, wfeng@ieee.org, songwang@cec.sc.edu

}

\usepackage{bibentry}

\begin{document}

	\maketitle
	
	\begin{abstract}
		Gait recognition is an important AI task, which has been progressed rapidly with the development of deep learning.
		However, existing learning based gait recognition methods mainly focus on the single domain, especially the constrained laboratory environment.
		In this paper, we study a new problem of unsupervised domain adaptive gait recognition (UDA-GR), that learns a gait identifier with supervised labels from the indoor scenes (source domain), and is applied to the outdoor wild scenes (target domain).
		For this purpose, we develop an uncertainty estimation and regularization based  UDA-GR method. Specifically, we investigate the characteristic of gaits in the indoor and outdoor scenes, for estimating the gait sample uncertainty, which is used in the unsupervised fine-tuning on the target domain to alleviate the noises of the pseudo labels.
		We also establish a new benchmark for the proposed problem, experimental results on which show the effectiveness of the proposed method.
		We will release the benchmark and source code in this work to the public.
	\end{abstract}

	\section{Introduction}
	
	Gait recognition is an important task and has many artificial intelligence (AI) applications, e.g., video surveillance.
	Existing gait recognition methods can be categorized into two types, i.e., the appearance-based~\cite{chai2022lagrange,fan2020gaitpart,lin2021gait} and model-based~\cite{liao2020model,li2021static,wang2022frame}, where the former uses the RGB image or binary silhouette sequence of a pedestrian as input and the latter uses the human pose sequence.
	For both of them, most state-of-the-art methods are developed based on the deep learning models, which are significantly depended on the training datasets.
	At present, the mainstream datasets for gait recognition are CASIA-B \cite{yu2006framework} and OUMVLP \cite{takemura2018multi}.
	These two datasets are both built in the laboratory environment, where the subjects all walk along a specified route and are captured by several fixed cameras. 
	The gait recognition algorithms trained under this setting is not very  practical in the real-world applications.
	Thus, an important problem is to extend the gait recognition from the laboratory environment (indoor) to the wild environment (outdoor).
	
\begin{figure}
	\centering
	\subfigure[]{
	\includegraphics[width=0.48\columnwidth]{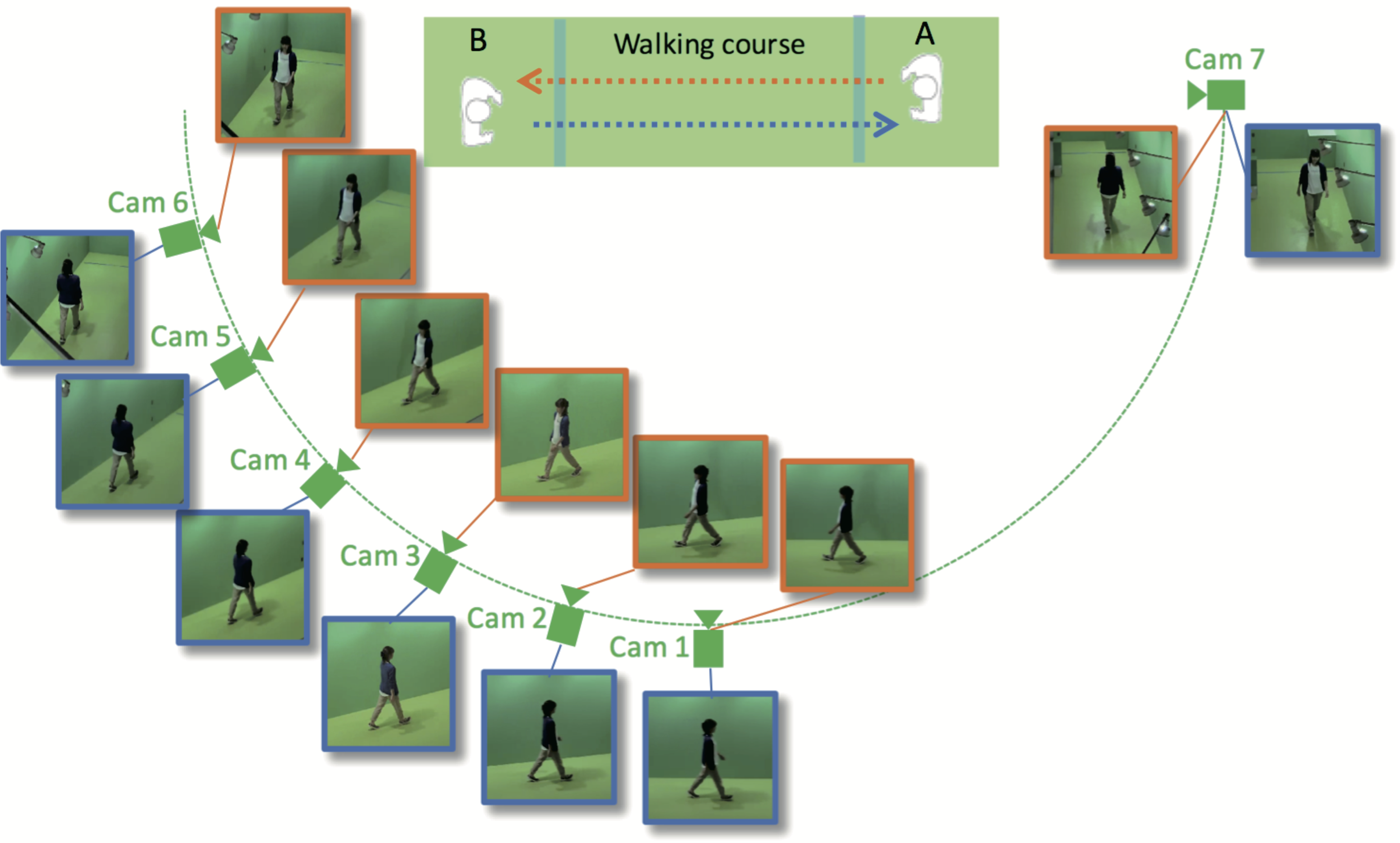}}
	\subfigure[]{
	\includegraphics[width=0.48\columnwidth]{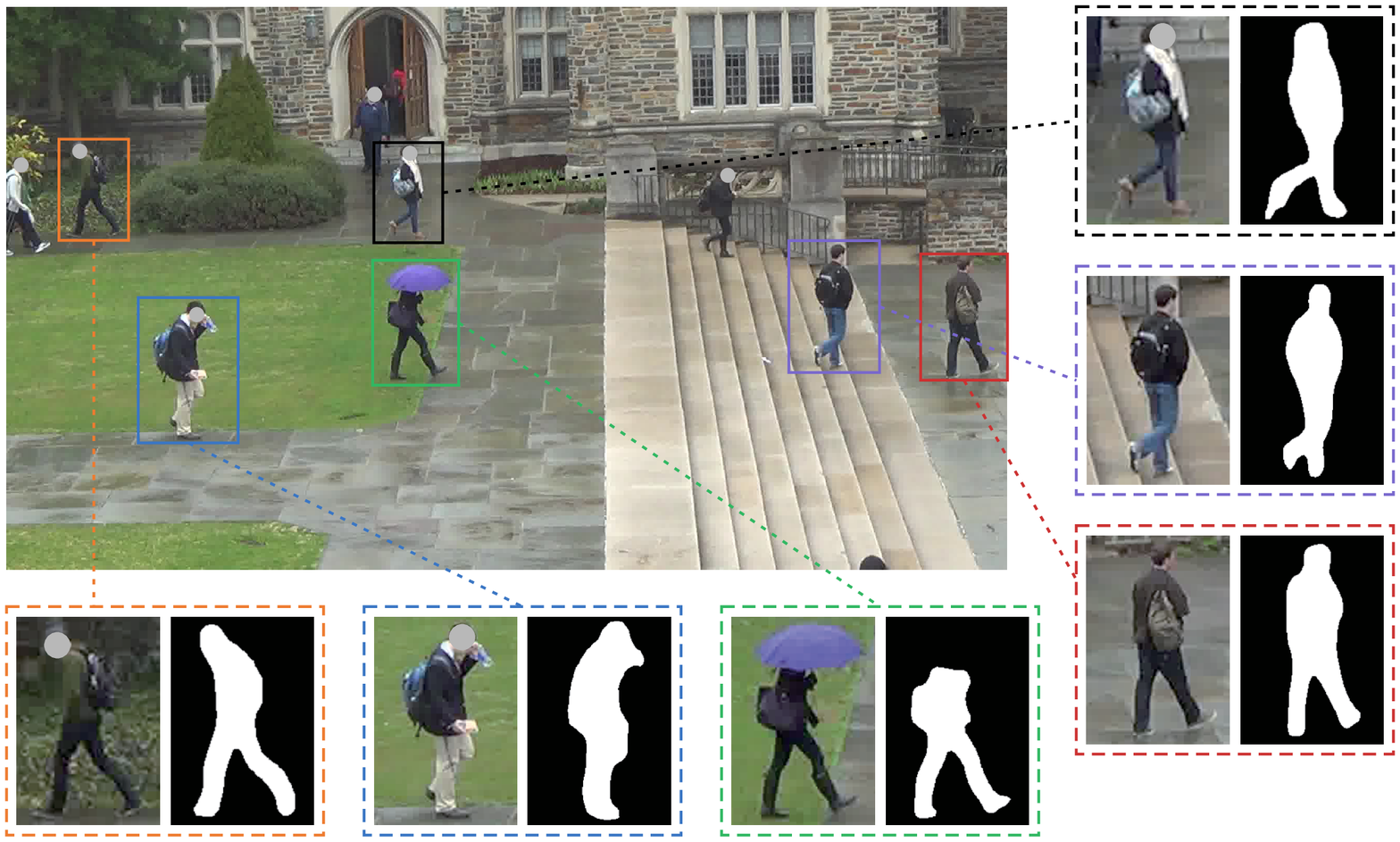}}
	\caption{A comparison between the gait recognition task the indoor (laboratory) and outdoor (wild) environments. (a) An illustration of the shooting condition of an indoor datasets in \citet{takemura2018multi}, where the cameras and walking courses of subjects are fixed. All samples have the same clean background. (b) A new outdoor gait recognition dataset built in this work, where the camera views and background are random and complex, belongings carrying and partial occlusions are also common.}
	\label{fig1} \vspace{-10pt}
	\end{figure}

	
	In this work, we propose to study a new problem of unsupervised domain adaptive gait recognition (UDA-GR), that is to learn a gait identifier with the supervised labels in the source domain and apply it to the target domain without supervised model training.
	We are more interested in the cross-domain gait recognition from `indoor' to `outdoor' as discussed above.
	However, given the large difference between the indoor and outdoor environments, the proposed UDA-GR is a very challenging problem.
	Specifically, for the indoor environment, i.e., the laboratory (Figure~\ref{fig1}(a)), multiple cameras are pre-installed with fixed locations and angles, subjects are repeatedly walking along a specific route, and the background is simple without change.
	This makes the gait sequence samples have no much additional discrimination, except for the different identifications. Also, the number of samples under different shooting angle and distance are uniformly distributed.
	In constraint, for the outdoor environment, e.g., the surveillance videos in the wild (Figure~\ref{fig1}(b)), the subject trajectories are random, and the background is complex.
	This way, the gait samples are various, and the number of samples under different shooting conditions are unevenly distributed.
	
	A similar problem to this work is UDA person Re-ID. As we observed, there are distinct styles in cross-domain person Re-ID samples, including backgrounds, lighting conditions, resolutions, etc. 
	A typical approach for this problem is to apply the GAN for performing the style transfer while keeping the person identity. Differently, for gait recognition datasets, such style gap does not exist, since the samples in gait recognition datasets are all binary silhouettes or human poses.
	{Also, different from UDA Re-ID where the source and target domain dataset are similar and exchangeable, UDA gait recognition we explore in this paper has a typical attribute, i.e., the source domain is from the constrained indoor scene while the target domain is from the complex outdoor scene.}
	
	In this work, we develop an uncertainty regularization based UDA-GR framework. Specifically, the framework is composed of the source domain training, target domain pseudo label generation and fine-tuning stages.	
	To handle the pseudo label noises under the large domain gap in this work, we apply a series gait transformations to simulate the outdoor gaits based on the indoor ones and use the feature discrepancy between the original and domain-transformed gaits to estimate the sample uncertainty.
	In the fine-tuning stage on the target domain,  the samples with higher uncertainty are inclined to be inactive given the noises of the pseudo labels.		
	The proposed simple and effective method achieves a promising result in UDA-GR compared to the existing gait recognition and UDA Re-ID methods. 
	
	The main contributions of this work are summarized as below.
	1) This is the first work to study the unsupervised domain adaptive gait recognition (UDA-GR) from the indoor to outdoor scenes, which is promising to promote the gait recognition to be more practical.
	2) We benchmark the UDA-GR problem by establishing four pairs of cross-domain datasets, in which we also build a new outdoor gait recognition dataset namely DukeGait.
	We further conduct the experiments on this benchmark to investigate the performance of the state-of-the-art gait recognition and UDA Re-ID methods.
	3) We develop a simple and effective uncertainty estimation and regularization based  UDA-GR method, by exploiting the sample uncertainty during the cross-domain learning. Experimental results show the effectiveness of the proposed method.
	We will release the benchmark and source code in this work to the public for promoting the researches on this new yet practical topic.

	\section{Related Work}
	
	\subsubsection{Gait Recognition.} Existing gait recognition methods can be mainly categorized into two types based on their raw input data, i.e., appearance-based \cite{huang2021context,fan2020gaitpart,lin2021gait} and model-based \cite{liao2020model,li2021static,wang2022frame}.
	Specifically, appearance-based methods generally extract discriminative features from the RGB image or binary silhouette sequence that represents the human gait. 
	Among them, some works \cite{wu2016comprehensive,zhang2019learning} compress the gait image sequence into a single template image, such as Gait Energy Image (GEI) \cite{han2005individual}, to treat gait recognition as a single-image-classification alike task. 
	Other methods take the gait sequence as input and model the temporal informations through temporal pooling \cite{chao2019gaitset,hou2020gait}, LSTM \cite{zhang2019cross,zhang2019gait} or 3D convolution \cite{lin2020gait}. 
	
	For another category, i.e., model-based methods, they take the skeleton data \cite{liao2017pose,liao2020model} or heatmaps \cite{feng2016learning} extracted by pose estimation algorithms \cite{cao2017realtime} as input and aim to model human body structure and walking patterns to achieve the gait recognition. 
	For such methods, CNN is commonly applied to extract spatial features and the LSTM is then used to extract the temporal features. Recently, inspired by the outstanding performance of GCN on many computer vision tasks, some works \cite{teepe2021gaitgraph,wang2022multi} have adopted the GCN for skeleton-based gait recognition.
	
	For thees two types of methods above, 
	with respect to the performance on existing datasets, there is still a gap between skeleton-based and appearance-based methods, due to the inaccurate pose estimation results and limited information contained in the skeleton data. 
	Therefore, we adopt the current mainstream appearance-based feature and explore its cross-domain adaption for gait recognition in this work.	
	
	\subsubsection{Gait Recognition Datasets.}
	Existing gait recognition methods are almost all developed based on the two most popular indoor datasets, i.e., CASIA-B \cite{yu2006framework} and OUMVLP \cite{takemura2018multi}. The two datasets are captured under predefined settings, that is, the subjects all walk along a fixed route and are captured by several fixed cameras. 
	Many other datasets are also collected in the laboratory environment, e.g., CMU MoBo \cite{gross2001cmu}, CASIA-A \cite{wang2003silhouette}, CASIA-C \cite{tan2006efficient}, CASIA-E \cite{song2022casia}, and TUM GAID \cite{hofmann2014tum}, etc.
	More recently, \citet{zhu2021gait} explored the outdoor gait recognition, and constructed the GREW dataset from natural videos in open systems.
	However, these datasets only focus on the single-domain gait recognition, where the styles of samples in the training set and testing set are similar.
	Note that, the indoor datasets for gait recognition are relatively easy to collect, but the algorithms based on them are not practical in the real world.
	The construction of the outdoor gait recognition datasets, especially for the identification annotation, is very laborious.
	So, it is significant to develop the unsupervised domain adaptation methods for gait recognition from indoor to outdoor.
	%
	%
	
	\begin{figure*}[t]
	\centering
	\includegraphics[width=0.875\textwidth]{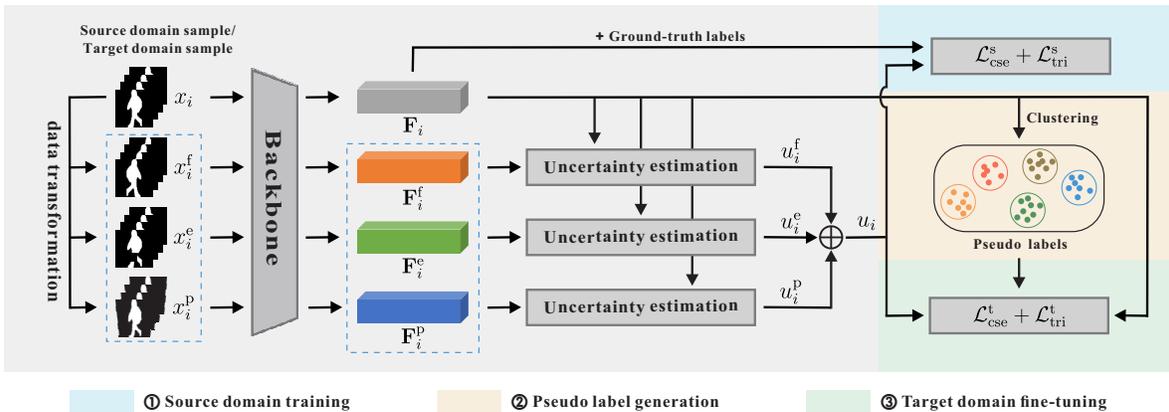}
	\caption{Framework of the proposed method for UDA-GR. In the source domain pre-training stage, we pre-train the model using the source domain samples with the given ground-truth labels. Then the clustering algorithm is applied on the features of unlabeled target domain samples to achieve pseudo labels. Finally, we fine-tune the model using the target domain data with the predicted pseudo labels. 
	We estimate the sample uncertainty based on the discrepancy between the original and transformed gaits, which is used to modify the cross-entropy and triplet losses, to guide both the network pre-training and fine-tuning.}
	\label{fig2} 	\vspace{-10pt}
	\end{figure*}
	
	\subsubsection{Unsupervised Domain Adaptation (UDA).} UDA aims to make the model trained on the source domain with label supervision to adapt a target domain without annotated labels.
	It has been widely studied for decades and achieved significant results on many tasks, e.g., image classification \cite{lee2019drop}, semantic segmentation \cite{toldo2020unsupervised}, and person re-identification \cite{zheng2021exploiting}. 
	A closely related topic to this work is the UDA person Re-ID, where the persons from the source domain and the target domain maybe completely different in the camera style, light condition, etc.
	The approaches for this problem have been widely studied~\cite{dai2021dual}. For example, the GAN-based methods \cite{deng2018image,wei2018person} consider to adopt GAN to generate the new training data from the source domain with the style of the target domain, to learn domain-invariant features from the style-transferred images.
	Another series of clustering-based methods \cite{zheng2021exploiting,fu2019self,dubourvieux2021unsupervised} generally generate pseudo labels for unlabeled target-domain data first, and then fine-tune the pre-trained model (on the source domain) using the data with pseudo labels. 
	Also, there are also other works \cite{zheng2021group,zhong2019invariance} propose to jointly train the network with the data from the source and target domains, using the similarities between the features of training samples and the instances in the memory bank.
	
	However, there are few works to systematically study UDA for the gait recognition problem. \citet{zheng2021trand} made the first attempt for unsupervised cross-domain gait recognition, and proposed a transferable neighborhood discovery framework to address this task.
	However, both the source domain and target domain in this work are from indoor datasets, where real-world scenes are not considered. 
	Different from that, we explore UDA gait recognition from the indoor dataset (source domain) to the outdoor dataset (target domain) in this paper, which is a valuable task with many theoretical investigations and practical applications.

	\section{Proposed Method}
	
	\subsection{Overview}
	The overall pipeline of the unsupervised domain adaptive gait recognition is illustrated in Figure~\ref{fig2}.
	UDA-GR takes a fully labeled source domain dataset $ \mathbb{D}^s=\{(x_{i}^\mathrm{s}, y_{i}^\mathrm{s})\vert _{i=1}^{N_s} \}$ and an unlabeled target domain dataset $ \mathbb{D}^{ \rm t}=\{(x_{j}^{t})\vert _{j=1}^{N_t} \}$ as input,  where $N_s$ and $N_t$ represent the number of gait sequences. For $i$-th gait sequence $x_{i}^\mathrm{s}$ in the source domain dataset, the corresponding label $y_{i}^\mathrm{s} \in \{1,2,\cdots,C^{ \rm t}\}$ is provided, where $C^{ \rm s}$ is the number of identities. For $j$-th gait sequence $x_{j}^\mathrm{t}$ in the target domain dataset, the corresponding identity label is unavailable.
	The goal of UDA-GR is to leverage both labeled source domain data and unlabeled target domain data to train a feature extractor and an identification classifier that can generalize well on the target domain.
	
	We develop a general UDA-GR framework including three main steps (Section~\ref{sec:UDA_framework}). First, given the gait sequence, we train the gait recognition network on the source domain with the given supervised labels.
	Then, on the target domain without supervision, we use the pre-trained feature extraction network together with a clustering algorithm for human identification, to get the pseudo labels.
	Finally, we fine-tune the network using the pseudo labels on the target domain.
	As discussed in Sections~\ref{sec:uncest}, there is a large domain gap between the source domain and the target domain. Also, in this work, the gait recognition on the target domain is always more difficult than that on source domain, which is different from the previous UDA person Re-ID task.
	This makes the pseudo labels on the target domain not very credible.
	To take full advantage of the pseudo labels, we propose a uncertainty-aware training strategy (Section~\ref{sec:uncer}).

	\subsection{UDA Gait Recognition Framework}
	\label{sec:UDA_framework}
	
	\textbf{Source domain training.}
	On the source domain, we apply an existing gait recognition model as the backbone network:
	\begin{equation}
	\label{eq:backbone}
	\mathbf{F}_i^\mathrm{s} = f(x_i^\mathrm{s}|\theta), \quad \mathbf{p}_i^\mathrm{s} = \varphi (\mathbf{F}_i^\mathrm{s}),
	\end{equation}
	where $f(\cdot |\theta)$ denotes the feature extraction operation and $\theta$ denotes the parameters of the model, $\mathbf{F}_i^\mathrm{s}$ denotes the extracted feature of the gait sequence $x_i^\mathrm{s}$. 
	We use $\varphi (\cdot )=\mathrm{softmax}(\mathrm{FC}(\cdot ))$ as the classifier function to map the feature into the probability vector $\mathbf{p}_i^\mathrm{s} \in \mathbb{R}^{C^\mathrm{s}}$, which encodes the probabilities of $x_i^\mathrm{s}$ belonging to each identity.
	
	We train the model $f$ and $\varphi$ using the labeled source domain data with the discrepancy between predicted and the ground-truth results
\begin{equation}
\mathcal{L}^\mathrm{s}= \mathbb{E}[\mathbf{p}_{i}^\mathrm{s} - \mathbf{y}_{i}^\mathrm{s}],
\end{equation}
	where $\mathbf{y}_{i}^\mathrm{s}$ is a one-hot vector, with $\mathbf{y}_{i}^\mathrm{s}(c) = 1$ if the index $c$  equals to the correct identify label $y_{i}^\mathrm{s}$, else $\mathbf{y}_{i}^\mathrm{s}(c) = 0$.	
	Specifically, the above loss, in this work, is implemented the combination of cross-entropy loss $\mathcal{L}_{\rm cse}^\mathrm{s}$
	\begin{equation}
	\label{eq:Lcse}
	\mathcal{L}_{\rm cse}^\mathrm{s}= - \sum\nolimits_{c} \mathbf{y}_{i}^\mathrm{s}(c) \log \mathbf{p}_i^\mathrm{s}(c).
	\end{equation}	
	and the triplet loss
	\begin{equation}
	\label{eq:Ltri}
	\mathcal{L}_{\rm tri}^\mathrm{s}= [d(\mathbf{F}_a^\mathrm{s}, \mathbf{F}_p^\mathrm{s}) - d(\mathbf{F}_a^\mathrm{s}, \mathbf{F}_n^\mathrm{s}) + m]_{+} ,
	\end{equation}
	where $\mathbf{F}_a^\mathrm{s}$, $\mathbf{F}_p^\mathrm{s}$, $\mathbf{F}_n^\mathrm{s}$ denote the feature of the anchor, positive, and negative sample, respectively. $d(\cdot, \cdot)$ denotes the Euclidean distance between the two feature vectors. $m$ denotes the margin of the triplet loss.
	
	\textbf{Pseudo label generation.}
	We feed the target domain data into the above pre-trained model and get the initial target domain features $\{\mathbf{F}_1^t, \mathbf{F}_2^t, \cdots , \mathbf{F}_{N_t}^t\}$. 
	We then apply a clustering algorithm ${ \rm Clu}(\cdot)$ on the features to compute clusters and predict pseudo labels for each target domain sample as
	\begin{equation}
	\{\hat{y}_1^\mathrm{t},\hat{y}_2^\mathrm{t},\cdots ,\hat{y}_{N_t}^\mathrm{t}\} = { \rm Clu}(\{\mathbf{F}_1^{ \rm t}, \mathbf{F}_2^{ \rm t}, \cdots , \mathbf{F}_{N_t}^{ \rm t}\}),
	\end{equation}
	where $\hat{y}_i^{ \rm t} \in \{1,2,\cdots,C^{ \rm t}\}$ denotes the identity pseudo label for each target sample $x_i^{ \rm t}$. $C^{ \rm t}$ denotes the number of clusters.
	
	\textbf{Target domain fine-tuning.}
	Finally, we fine-tune the model using the target domain data with the predicted pseudo labels. The loss function is defined as 
	\begin{equation}
	\label{eq:losst}
	\mathcal{L}^\mathrm{t}= \mathbb{E}[\mathbf{p}_{i}^\mathrm{t} - \mathbf{\hat{y}}_{i}^\mathrm{t}] + \mathbb{E}[\mathbf{\hat{y}}_{i}^\mathrm{t}-\mathbf{y}_{i}^\mathrm{t}],
	\end{equation}
	which is combined of the discrepancy between the prediction and the pseudo label (first term), and the discrepancy between the pseudo label and the ground-truth label (second term).
	The second term is a constant since the pseudo labels are fixed during the fine-tuning.
	Therefore, we actually optimize the first term in this stage.
	We also use the cross-entropy loss and triplet loss here. 
	The pseudo label generation and model fine-tuning with pseudo labels are alternated until the model converges on the target domain.

	\subsection{Uncertainty Estimation}	
	\label{sec:uncest}
	For the UDA-GR problem, there are significant distribution gap and scene difference between the source domain dataset (over 100 sequences for each subject in CASIA-B dataset \cite{yu2006framework}, indoor scene, pre-fixed route, and horizontal-view camera) and target domain dataset (only average 5 sequences for each subject in GREW dataset \cite{zhu2021gait}, outdoor scene, free route, and downward sloping top-view camera).
	This makes the generalization performance of the above UDA-GR framework not strong enough. 
	This is because,
	when fine-tuning on the target domain, the generated pseudo labels contain unavoidable noise, which undoubtedly affects the subsequent feature learning. 
	To handle this problem, we propose an uncertainty-aware training strategy to alleviate 
	the negative influence of noisy pseudo labels on the target domain, based on the estimated uncertainty of each input sequence.
	
	\begin{figure}[t]
		\centering
		\includegraphics[width=0.8\columnwidth]{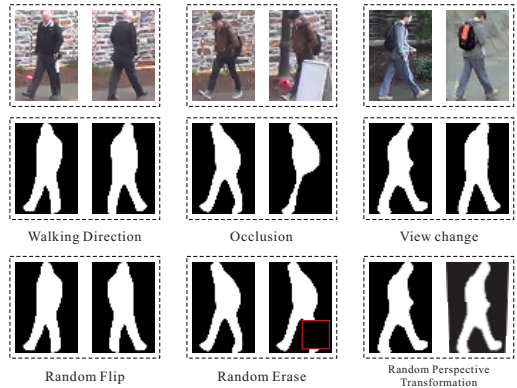} 
		\caption{Examples of common disturbed conditions, i.e., walking direction change, occlusion, and camera view change, in outdoor dataset. Each pair of RGB images (the first row) and the corresponding silhouettes (the second row) contain the same subject, and random flip, random erase, and random perspective transformation (the third row) are designed to simulate these conditions.}
		\label{fig3} 	\vspace{-20pt}
	\end{figure}

	The uncertainty can be naturally formulated as 
	\begin{equation}
	\mathrm{Unc}(x_{i}^\mathrm{t}) = \mathcal{D}(\varphi(\mathbf{F}_i^\mathrm{t}), \mathbf{y}_{i}^\mathrm{t}),
	\end{equation}
	which denotes the discrepancy of the prediction as defined in Eq.~(\ref{eq:backbone}) and the label $\mathbf{y}_{i}^\mathrm{t}$.	
	On the target domain, the label $\mathbf{y}_{i}^\mathrm{t}$ is not available. A naive way is to utilize the pseudo label $\mathbf{\hat{y}}_{i}^\mathrm{t}$ to replace it. The uncertainty are then approximated as
	\begin{equation}
	\mathrm{Unc}(x_{i}^\mathrm{t}) \approx  \mathcal{D}(\varphi(\mathbf{F}_i^\mathrm{t}), \mathbf{\hat{y}}_{i}^\mathrm{t}),
	\end{equation}
	However, we can see from Eq.~(\ref{eq:losst}) that the predicted $\mathbf{p}_{i}^\mathrm{t}$ is compelled to close the pseudo labels $\mathbf{\hat{y}}_{i}^\mathrm{t}$.
	In this work, we adopt another approximation as
	\begin{equation}
	\label{eq:unc3}
	\mathrm{Unc}(x_{i}^\mathrm{t}) \approx  \mathcal{D}(\varphi(\mathbf{F}_i^\mathrm{t}), \varphi(\mathbf{\tilde{F}}_i^\mathrm{t})),
	\end{equation}
	which is specifically implemented as below.
	
	As shown in Figure~\ref{fig3} (the first two rows), in gait recognition datasets, especially outdoor datasets with free-setting, some common and inevitable conditions, e.g., random walking direction, mutual occlusions, and camera view change, generally bring the changes to the silhouettes of the same subject. 
	The gait recognition methods strive to handle such changes and learn robust feature representations from the samples of the same person under various conditions.
	This way, in this work, we estimate the uncertainty for the pseudo labels based on the consistency between feature representations under different conditions of the same subject.
	
	Specifically, we design three gait sequence transformation manners, including random flip, random erase, and random perspective transformation, to simulate walking direction change, occlusion, and camera view change, respectively.
	The uncertainty for each sample is estimated based on the consistency between the feature vectors of the original gait sequence and the transformed one.
	Given a gait sequence sample $x_{i}^\mathrm{t}$, we obtain three types of transformed samples $x_i^{\rm f}$, $x_i^{\rm e}$, $x_i^{\rm p}$ through the above gait transformation manners, i.e., random flip, random erase, and random perspective transformation, as shown in Figure~\ref{fig3} (the third row). 
	The transformed samples are fed into the model to get the corresponding feature representations $\mathbf{F}_i^{\rm f}$, $\mathbf{F}_i^{\rm e}$, $\mathbf{F}_i^{\rm p}$ as $\mathbf{\tilde{F}}_i^\mathrm{t}$ in Eq.~(\ref{eq:unc3}). 

	Note that, in Eq.~(\ref{eq:unc3}), the classifier $\varphi (\cdot )$ is changed with different output dimensions, since 1) the number of identifies for training differs between the source and target domains, 2) the number of clusters during clustering on the target domain is changed. 
	Therefore, we replace the predicted probability with the feature before the classification layer and get the uncertainty estimation as
	$u_i^{\rm t} = \mathcal{D}(\mathbf{F}_i^\mathrm{t}, \mathbf{\tilde{F}}_i^\mathrm{t})$.
	$
	$

	\subsection{Uncertainty-regularized Learning}
	\label{sec:uncer}
		
	As discussed in the UDA-GR framework, we apply the modified cross-entropy loss and triplet loss as the total loss for network learning.
    We then discuss about how to use the uncertainty in the framework.

	\textbf{Uncertainty-regularized target domain fine-tuning.}
	On the target domain, we use pseudo labels for model fine-tuning. Therefore, the higher uncertainty of a sample, the more noise it has, that is, the more likely it is assigned an incorrect pseudo label in the pseudo label generation stage.
	With the uncertainty, we propose to adjust the weight of each sample in the model fine-tuning phase. 
	For an input sample with high uncertainty (noisy sample), we will reduce its penalty weight in the loss function, while for a sample with lower uncertainty, the weight will be increased. 
	
	Specifically, as in the definition of uncertainty-guided re-weighted loss in \citet{arnab2020uncertainty}, based on Eq.~(\ref{eq:Lcse}), we modify the cross-entropy loss on the target domain as
	\begin{equation}
	\mathcal{L}_{\rm cse}^{\rm t}=- (1 - u_i^{\rm t})  \sum\nolimits_{c} \mathbf{y}_{i}^\mathrm{s}(c) \log \mathbf{p}_i^\mathrm{s}(c)  - \log (1 - u_i^{\rm t}),
	\end{equation}
	where $u_i^{\rm t}$ denotes the uncertainty of $x_{i}^\mathrm{t}$, whose value is in the range of [0,1].
	For the triplet loss, based on in Eq.~(\ref{eq:Ltri}),  we adopt the policy in \citet{zheng2021exploiting} and calculate the pair credibility through
	\begin{equation}
	\label{eq:triw2}
	\omega _{\rm ap}^\mathrm{t} = (1-u_a) + (1-u_p), \  \omega _{\rm an}^\mathrm{t} = (1-u_a) + (1-u_n),
	\end{equation}	
	where $u_a$, $u_p$, $u_n$ denote the estimated uncertainty of the anchor sample, positive sample and negative sample in a triplet, respectively. $\omega _{\rm ap}^t$ denotes the pair credibility of the positive sample pair, and $\omega _{\rm an}^t$ for the negative sample pair.
	The modified triplet loss on the target domain is defined as
	\begin{equation}
	\label{eq:tri}
	\begin{aligned}
	\mathcal{L}_{\rm tri}^{\rm t}=-[\omega _{\rm ap}^{\rm t} d(\mathbf{F}_a^{\rm t}, \mathbf{F}_p^{\rm t}) - \omega _{\rm an}^{\rm t} d(\mathbf{F}_a^{\rm t}, \mathbf{F}_n^{\rm t}) + m]_{+},
	\end{aligned}
	\end{equation} 
	where the lower pair credibility (the higher uncertainty) of a sample pair, the lower penalty weight of that in the loss, and thus the lower weight it makes to the model optimization.
	
	\textbf{Uncertainty-aware source domain training.}
	On the source domain training, 	we use the ground-truth identity labels for model training. 
	This way, we consider to make the original feature $\mathbf{F}_i^\mathrm{s}$ and the transformed one $\mathbf{\tilde{F}}_i^\mathrm{s}$ as close as possible on the source domain training.
	This can be regarded as the data augmentation that makes the data on source domain be similar to that on target domain.
	Also, this can make the difference of $\mathbf{F}_i^\mathrm{t}$ and the transformed one $\mathbf{\tilde{F}}_i^\mathrm{t}$ better depict the uncertainty on the target domain. 
	
	Specifically, we modify the cross-entropy loss on the source domain as
	\begin{equation}
	\mathcal{L}_{\rm cse}^{\rm s}=- u_i^\mathrm{s} \sum\nolimits_{c} \mathbf{y}_{i}^\mathrm{s}(c) \log \mathbf{p}_i^\mathrm{s}(c)  - \log u_i^\mathrm{s},
	\end{equation}
	where $u_i^{\rm s}$ denotes the uncertainty of $x_{i}^\mathrm{s}$ on source domain, i.e., the difference between $\mathbf{F}_i^\mathrm{s}$ and $\mathbf{\tilde{F}}_i^\mathrm{s}$. Intuitively, for the input samples with high uncertainty (hard sample), we will increase its weight in the loss function for the model to give more attention during training. 
	Similarly, for the triplet loss, we calculate the pair credibility in Eq.~(\ref{eq:triw2}) as 	
	\begin{equation}
	\label{eq:triw}
	\begin{aligned}
	\omega _{\rm ap}^s = u_a + u_p, \ \omega _{\rm an}^s = u_a + u_n.
	\end{aligned}
	\end{equation}	
	

	\subsection{Implementation Details}
	We use GaitGL \cite{lin2021gait} as our backbone network and first train it on the source domain dataset by following the original training strategy in \citet{lin2021gait}.
	During the fine-tuning stage, we apply the DBSCAN clustering algorithm~\cite{ester1996density} to generate pseudo labels for the target domain data. 
	The function $\mathcal{D}$ defined in Section~\ref{sec:uncest} is implemented by the cosine similarity on the source domain, and the learnable fully-connected layer on the target domain.
	All images are resized to $64\times 44$.
	The length of input gait sequences is set to 30 during training, and the whole gait sequences are fed into the model to extract features during test.
	We use the Adam algorithm to optimize the model in all experiments.

	\begin{table*}[]
		\caption{Comparison with the state-of-the-art gait recognition and UDA-Re-ID methods on UDA-GR benchmark using the metrics of mAP, Rank 1, Rank 5 and Rank 10 (\%).} 	\vspace{-10pt}
		\label{tab1}
		\centering
		\small
		\begin{spacing}{1.05}
			\begin{tabular}{lccccccccc}
				\hline
				\multicolumn{1}{c}{\multirow{3}{*}{Methods}} & \multicolumn{1}{c}{\multirow{3}{*}{Reference}} & \multicolumn{4}{c}{UDAGR-C2G} & \multicolumn{4}{c}{UDAGR-C2D} \\ \cline{3-10} 
				\multicolumn{1}{c}{\multirow{2}{*}{}}  & \multicolumn{1}{c}{\multirow{2}{*}{}}  & \multicolumn{4}{c}{CASIA-B$\rightarrow $GREW} & \multicolumn{4}{c}{CASIA-B$\rightarrow $DukeGait} \\ \cline{3-10}
				\multicolumn{1}{c}{} & \multicolumn{1}{c}{} & mAP & R1 & R5 & R10 & mAP & R1 & R5 & R10 \\ \hline
				UNRN \cite{zheng2021exploiting} & AAAI 2021 & 11.5 & 20.2 & 32.1 & 39.0 & 7.9 & 21.7 & 34.8 & 40.7 \\
				\rowcolor{mygray}
				IDM \cite{dai2021idm} & ICCV 2021 & 9.8 & 18.1 & 28.4 & 33.6 & 7.9 & 21.3 & 34.6 & 41.8 \\
				SECRET \cite{he2022secret} & AAAI 2022 & 6.5 & 12.3 & 19.6 & 24.8 & 5.4 & 12.6 & 23.4 & 29.2 \\ \hline
				\rowcolor{mygray}
				GaitSet \cite{chao2019gaitset} & AAAI 2019 & 14.0 & 24.5 & 37.4 & 45.1 & 8.4 & 24.3 & 39.5 & 46.7 \\
				GaitPart \cite{fan2020gaitpart} & CVPR 2020 & 14.2 & 25.5 & 39.0 & 45.5 & 8.5 & 25.4 & 39.8 & 46.6 \\
				\rowcolor{mygray}
				GaitGL \cite{lin2021gait} & ICCV 2021 & 15.7 & 30.0 & 40.5 & 46.6 & 13.0 & 40.7 & 56.9 & 62.2 \\ \hline
				Ours & This paper & 19.6 & 37.4 & 49.2 & 53.8 & 17.1 & 50.6 & 64.7 & 69.8  \\ \hline
			\end{tabular}
		\end{spacing}
	\end{table*}
	
	\begin{table*}[]	\vspace{-10pt}
		\centering
		\small
		\begin{spacing}{1.05}
			\begin{tabular}{lccccccccc}
				\hline
				\multicolumn{1}{c}{\multirow{3}{*}{Methods}} & \multicolumn{1}{c}{\multirow{3}{*}{Reference}} & \multicolumn{4}{c}{UDAGR-O2G} & \multicolumn{4}{c}{UDAGR-O2D} \\ \cline{3-10} 
				\multicolumn{1}{c}{\multirow{2}{*}{}}  & \multicolumn{1}{c}{\multirow{2}{*}{}} & \multicolumn{4}{c}{OUMVLP$\rightarrow $GREW} & \multicolumn{4}{c}{OUMVLP$\rightarrow $DukeGait} \\ \cline{3-10}
				\multicolumn{1}{c}{} & \multicolumn{1}{c}{} & mAP & R1 & R5 & R10 & mAP & R1 & R5 & R10 \\ \hline
				UNRN \cite{zheng2021exploiting} & AAAI 2021 & 7.3 & 12.7 & 22.6 & 27.9 & 4.8 & 11.8 & 22.5 & 27.4 \\
				\rowcolor{mygray}
				IDM \cite{dai2021idm} & ICCV 2021 & 11.1 & 19.8 & 30.5 & 36.6 & 7.7 & 20.5 & 34.0 & 41.3 \\
				SECRET \cite{he2022secret} & AAAI 2022 & 11.2 & 20.1 & 30.9 & 37.0 & 7.0 & 18.2 & 29.3 & 35.9 \\ \hline
				\rowcolor{mygray}
				GaitSet \cite{chao2019gaitset} & AAAI 2019 & 17.6 & 30.2 & 46.5 & 52.7 & 7.1 & 22.1 & 32.7 & 38.5 \\
				GaitPart \cite{fan2020gaitpart} & CVPR 2020 & 20.1 & 34.1 & 50.7 & 56.4 & 10.0 & 30.0 & 42.6 & 50.1 \\
				\rowcolor{mygray}
				GaitGL \cite{lin2021gait} & ICCV 2021 & 24.5 & 42.1 & 56.2 & 63.0 & 13.4 & 40.8 & 53.8 & 59.0 \\ \hline
				Ours & This paper & 28.5 & 47.0 & 60.2 & 65.7 & 17.6 & 51.2 & 64.6 & 70.9  \\ \hline
			\end{tabular}
		\end{spacing} 	\vspace{-10pt}
	\end{table*}
	
	\section{UDA-GR Benchmark}
	
	We do not find the available benchmarks for gait recognition across the indoor and outdoor domain.
	Therefore, we aim to establish the benchmark for this problem.
	Specifically, for the indoor gait recognition dataset, we select two mainstream datasets, i.e., CASIA-B, OUMVLP.
	The outdoor gait recognition dataset is not common, an available one is the recent GREW dataset.
	In this work, we also collect a new outdoor gait dataset namely DukeGait based on previous work.\\
	$\bullet$ CASIA-B \cite{yu2006framework} is an indoor dataset, which consists of 124 subjects captured from 11 views ($0^\circ, 18^\circ, 36^\circ, \cdots, 180^\circ$). Each subject under each view consists of 10 gait sequences under different conditions, i.e., six normal walking (NM), two carrying bag (BG), and two wearing coat (CL). Following \citet{wu2016comprehensive}, 74 subjects are used for training and the rest 50 for testing.
	\\
	$\bullet$ OUMVLP \cite{takemura2018multi} is also an indoor dataset, which consists of 10,307 subjects (5,114 males and 5,193 females with various ages) captured from 14 view angles ($0^\circ, 15^\circ, 30^\circ, \cdots, 90^\circ, 180^\circ, 195^\circ, 210^\circ, \cdots, 270^\circ$). Each subject under each view consists of 2 gait sequences. Following the official partition setting, 5,153 subjects are used for training and the rest 5,154 subjects for testing.
	\\
	$\bullet$ GREW \cite{zhu2021gait} is the first large-scale dataset for gait recognition in the wild, which consists of 26,345 subjects and 128,671 sequences, collected from 882 cameras in large public areas.
	In the official partition setting, 102,887 sequences of 20,000 subjects are used as the training set.
	Considering the comparable data scale with other datasets, from the GREW training set, we select the first 2,000 (00001-02000) subjects with 10,263 sequences as our training set, and the last 2,000 (18001-20000) subjects with 10,190 sequences for testing.
	\\
	$\bullet$ DukeGait is a new outdoor dataset we created in this work. It is built based on the Duke Multi-Target Multi-Camera Tracking (DukeMTMC) dataset  \cite{ristani2016performance}, whose videos come from the surveillance video footage taken on Duke University's campus. 
	DukeMTMC is used for human tracking task, which uses 8 non-overlapped cameras located at different sites in the campus for recording the pedestrians for 85 minutes. It  provides the human bounding boxes with unified IDs as the annotations.
	To obtain the gait recognition dataset, we first capture the trajectory of each pedestrian in a camera, which is then splitted into short tracks (about 200 frames).
	Besides, we also adopt an instance segmentation algorithm, i.e., HTC \cite{chen2019hybrid}, to extract silhouettes from original RGB sequence. 
	Finally, we use the bounding boxes in each short track to crop out silhouette sequences of each subject and get the gait sequence.
	The ID of each gait sequence is inherited from that of the original trajectory.
	In total, the proposed DukeGait consists of 1,819 subjects and 16,878 gait sequences, of which 1,000 subjects are used for training and the rest 819 subjects for testing.
	
	Based on the above datasets, we build the UDAGR dataset containing four protocols, including UDAGR-C2G, UDAGR-C2D, UDAGR-O2G, UDAGR-O2D, each of which is composed of an indoor dataset, i.e., CASIA-B (C), OUMVLP (O) and an outdoor dataset, i.e., GREW (G), DukeGait (D).

	\textbf{Metrics.} We adopt the classical gait recognition metrics, including the Rank-1/5/10 (R1/R5/R10) of CMC (Cumulative Match Characteristic) curve and the mAP (mean average precision) score as the evaluation protocols.
	
		\begin{table*}[]
		\caption{Ablation studies on UDA-GR benchmark for verifying the effectiveness of the components in our method (\%).} 	\vspace{-10pt}
		\centering
		\small
		\label{tab:abla}
		\begin{spacing}{1.05}
			\begin{tabular}{lcccccccc}
				\hline
				\multicolumn{1}{c}{\multirow{3}{*}{Methods}}& \multicolumn{4}{c}{UDAGR-C2G} & \multicolumn{4}{c}{UDAGR-C2D} \\ \cline{2-9} 
				\multicolumn{1}{c}{\multirow{3}{*}{}}  & \multicolumn{4}{c}{CASIA-B$\rightarrow $GREW} & \multicolumn{4}{c}{CASIA-B$\rightarrow $DukeGait}  \\ \cline{2-9} 
				\multicolumn{1}{c}{} & mAP & R1 & R5 & R10 & mAP & R1 & R5 & R10 \\ \hline
				Supervised learning on $\mathcal{T}$ (Oracle) & 35.7 & 54.0 & 68.3 & 74.4 & 21.3 & 55.4 & 69.9 & 75.6 \\
				\rowcolor{mygray}
				Direct testing on $\mathcal{T}$ (Baseline) & 15.7 & 30.0 & 40.5 & 46.6 & 13.0 & 40.7 & 56.9 & 62.2  \\
				UDA-GR w K-means & 13.0 & 25.6 & 37.1 & 42.1 & 10.3 & 34.0 & 50.3 & 56.1 \\
				\rowcolor{mygray}
				UDA-GR w DBSCAN & 14.1 & 27.8 & 38.4 & 43.7 & 13.3 & 43.6 & 56.1 & 61.5 \\
				UDA-GR w Unc. on $\mathcal{S}$ & 16.8 & 31.6 & 43.5 & 49.0 & 13.4 & 40.8 & 55.9 & 61.7 \\
				\rowcolor{mygray}
				UDA-GR w Unc. on $\mathcal{S+T}$ (Ours) & 19.6 & 37.4 & 49.2 & 53.8 & 17.1 & 50.6 & 64.7 & 69.8 \\
				\hline
			\end{tabular}
		\end{spacing}
	\end{table*}
	
	\begin{table*}[] 	\vspace{-10pt}
		\centering
		\small
		\begin{spacing}{1.05}
			\begin{tabular}{lcccccccc}
				\hline
				\multicolumn{1}{c}{\multirow{3}{*}{Methods}}& \multicolumn{4}{c}{UDAGR-O2G} & \multicolumn{4}{c}{UDAGR-O2D} \\ \cline{2-9} 
				\multicolumn{1}{c}{\multirow{3}{*}{}} & \multicolumn{4}{c}{OUMVLP$\rightarrow $GREW} & \multicolumn{4}{c}{OUMVLP$\rightarrow $DukeGait} \\ \cline{2-9} 
				\multicolumn{1}{c}{} & mAP & R1 & R5 & R10 & mAP & R1 & R5 & R10 \\ \hline
				Supervised learning on $\mathcal{T}$ (Oracle) & 45.6 & 64.0 & 77.3 & 81.4 & 28.5 & 69.0 & 80.2 & 84.1 \\
				\rowcolor{mygray}
				Direct testing on $\mathcal{T}$ (Baseline) & 24.5 & 42.1 & 56.2 & 63.0 & 13.4 & 40.8 & 53.8 & 59.0  \\
				UDA-GR w K-means & 18.4 & 32.6 & 47.1 & 53.5 & 9.8 & 31.9 & 47.6 & 53.0 \\
				\rowcolor{mygray}
				UDA-GR w DBSCAN & 19.4 & 35.9 & 47.6 & 53.7 & 13.0 & 43.0 & 57.3 & 62.4 \\
				UDA-GR w Unc. on $\mathcal{S}$ & 22.3 & 37.9 & 53.3 & 59.3 & 13.2 & 39.9 & 53.2 & 57.7 \\
				\rowcolor{mygray}
				UDA-GR w Unc. on $\mathcal{S+T}$ (Ours) & 28.5 & 47.0 & 60.2 & 65.7 & 17.6 & 51.2 & 64.6 & 70.9 \\ \hline
			\end{tabular}
		\end{spacing} 	\vspace{-10pt}
	\end{table*}

	\section{Experimental Results}
	
	\subsection{Comparison with State-of-the-Art Methods}
	\textbf{Comparison methods.}
	Since this is the first work to explore UDA-GR from the indoor to outdoor scenes, there is no method directly developed for this problem. 
	We compare our method with the state-of-the-art gait recognition and UDA ReID methods. 
	For the gait recognition methods, we select GaitSet \cite{chao2019gaitset}, GaitPart \cite{fan2020gaitpart}, GaitGL \cite{lin2021gait} for comparison. Different from using the same domain for training and test in the original setting, we train the networks with the source domain, i.e., the indoor datasets, and  perform the evaluation on the target domain, i.e., the outdoor datasets in the proposed UDA-GR benchmark.
	For the UDA ReID methods, we select UNRN \cite{zheng2021exploiting}, IDM \cite{dai2021idm}, SECRET \cite{he2022secret} for comparison. Note that, these method all take a single image as input, therefore we compress the gait sequence into a single gait template GEI, and feed it to the networks for training and test.

	\textbf{Results.}
	As shown in Table~\ref{tab1} (top 3 rows), the UDA Re-ID methods can not work well when applied directly to the UDA-GR task. 
	This is because 1) These methods designed for image Re-ID extract features from a single image, losing the temporal information that is significant for gait recognition. 
	2) These cross-domain methods are specifically designed for the UDA Re-ID task. Compared to UDA Re-ID, the UDA-GR task has a larger domain gap, e.g., more difference with respect to the data distribution between the source and target domains, which is not considered in these methods.
	Although the gait recognition methods achieve a slightly higher performance (middle 3 rows), the results are still not satisfactory. This is because 1) These methods are developed for the single-domain gait recognition. 2) They consider only the indoor scenes.
	In addition, we can see that the models pre-trained on OUMVLP generally perform better than those pre-trained on CASIA-B, which indicates that the larger source domain datasets can better help with the cross-domain gait recognition task.
	As shown in the last row, we can see that our method significantly outperforms all the state-of-the-art gait recognition and UDA Re-ID methods for all cross-domain settings on the UDA-GR dataset.
	
	\subsection{Ablation Study}
	
	$\bullet$ Supervised learning on $\mathcal{T}$: After the training on source domain, we fine-tune the network on the target domain $\mathcal{T}$ using the supervised labels. This result generated by the training with supervision can be regarded as the oracle of the proposed unsupervised method. \\
	$\bullet$ Direct testing on $\mathcal{T}$: We directly apply the network trained on source domain to the target domain test. This can be regarded as the baseline of this work.\\
	$\bullet$ UDA-GR w K-means / DBSCAN: We apply the proposed UDA gait recognition framework in Section~\ref{sec:UDA_framework} with the clustering algorithms of K-means~\cite{lloyd1982least} and DBSCAN~\cite{ester1996density}, respectively.  \\
	$\bullet$ UDA-GR w Unc. on $\mathcal{S}$ / $\mathcal{S+T}$: Based on the UDA-GR framework, we add the uncertainty-aware training strategy discussed in Section~\ref{sec:uncer} on the source domain ($\mathcal{S}$) training or also on the target domain fine-tuning ($\mathcal{S+T}$).\\
	From Table~\ref{tab:abla}, we can first see that, compared to the direct test (baseline) method, the pseudo label based UDA-GR framework provides even poorer results. 
	Note that, the clustering-based pseudo label generation is generally used in previous UDA problems and commonly generate a better performance, which, however, is not very effective here. This is because the pseudo labels are with more noises given the difficult target-domain samples and large domain gap in this problem.
	With the proposed uncertainty strategy,
	we can see that the integration of uncertainty only on source-domain training (with direct test on the target domain) partly improve the performance of the baseline on CASIA related datasets but is not effective on OUMVLP related datasets.
	This is because the uncertainty-aware training on source-domain is used to help the target-domain fine-tuning with uncertainties.
	But the independent usage of it is not very effective.
	As shown in the last row, with the uncertainty strategy also equipped to the target-domain fine-tuning in our UDA-GR framework, the performance is further improved on all four pairs of datasets with a large margin of 3.9\%, 4.1\%, 4.0\%, 4.2\% on mAP score, respectively.
	
		\begin{figure}[t]
		\centering
		\includegraphics[width=1\columnwidth]{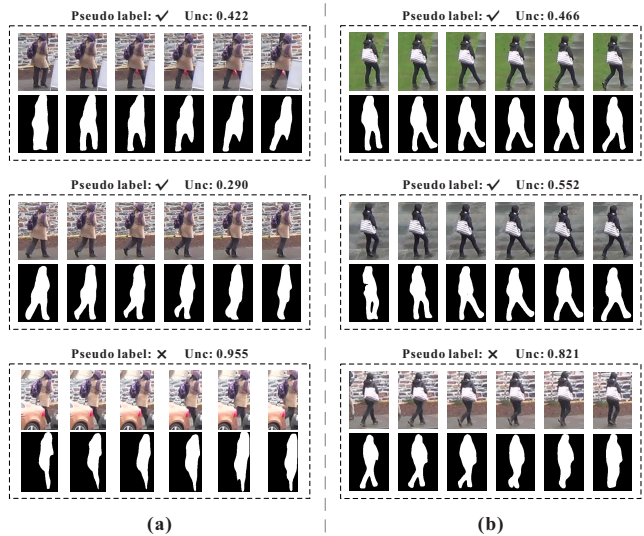}  \vspace{-15pt}
		\caption{Examples of pseudo label and uncertainty prediction results on DukeGait. 
		The uncertainties are normalized by min-max scaling.}
		\label{fig4} \vspace{-15pt}
	\end{figure}

	\subsection{Experimental Analysis}
	To further verify the effectiveness of the proposed uncertainty, we provide some illustrations.
	Figure~\ref{fig4} shows six examples of pseudo label and uncertainty prediction results from DukeGait.
	The left three samples (a) show the same subject from the same camera (view). 
	Compared to the sample with slight occlusion or without occlusion, the last one with severe occlusion with the wrong pseudo label prediction has a higher uncertainty.
	The right samples (b) show the same subject from three different cameras (views).
	We observe that compared to the top two samples, the last one with wrong pseudo label and higher uncertainty, has a less discriminative gait sequence, especially in terms of opening/closing legs.
	This demonstrates that our method can actually predict the sample uncertainties on the target domain, therefore assign discriminative weights to the pseudo labels.	
	
	\section{Conclusion}
	
	In this paper, we have studied a new challenging and practical problem of unsupervised domain adaption gait recognition (UDA-GR) from the indoor to outdoor scenes. For this problem, we propose a pseudo label fine-tuning based UDA-GR framework, which is further improved with an  uncertainty-regularized learning strategy to handle problems of label noise in target-domain fine-tuning.
	We also build a new benchmark for the proposed UDA-GR problem with a newly collected outdoor gait recognition dataset and other three mainstream public datasets.
	With the above effort, we hope to build the basics for the research of this new yet promising topic. 
	
	\bibliographystyle{aaai23}
	\bibliography{UDAGR}
	

\end{document}